\documentclass[11pt]{article}
\usepackage{tikz}
\usepackage{float}
\usepackage{amsmath}
\usepackage{hyperref}
\usepackage{algorithm}
\usepackage{algpseudocode}
\usepackage{multirow}

\title{\textbf{Investigation Into The Effectiveness Of Long Short Term Memory Networks For Stock Price Prediction}}
\author{Hengjian Jia \\ Colyton Grammar School Year 12 Student \\ E-mail: \href{mailto:henryjia18@gmail.com}{henryjia18@gmail.com}}
\date{}

\begin{document}
\maketitle

\section{Abstract}

The effectiveness of long short term memory networks trained by backpropagation through time for stock price prediction is explored in this paper. A range of different architecture LSTM networks are constructed trained and tested.

\section{Introduction}

Stock data and prices are a form of time series data. Classical macroeconomic and business based methods methods are traditionally used to tackle the problem of stock market prediction. These methods rely on human observation of patterns and corporate information\cite{TA1}.

Deep multilayer perceptrons \cite{MLP1} and convolutional neural networks \cite{CNN1} have also been used to tackle the problem of stock price prediction. However, these methods have limited capability for temporal memory which can be provided through a fixed sized sliding window for predicting future stock prices as a function of historical prices.

Recurrent neural networks have a cycle which feeds activations from the previous time step back in as an input and influences the activations of the current time step. Therefore the activations create an internal state. This in theory can store temporal information for a dynamic indefinite number of time steps in contrast to the fixed number of time steps of feed forward networks. LSTMs are a specific type of recurrent neural network which overcomes some of the problems of recurrent networks.\cite{LSTM2}.

\section{Long Short Term Memory}

LSTM hidden layers are made up of cells with sigmoidal input, output and forget gates. This allows the network to learn when to forget, take input and output. The LSTM cell has an internal state which is updated based on the previous activations of the layer and inputs through connections to the previous layer and self connections\cite{LSTM1}.

\begin{figure}[H]
\centering
\begin{tikzpicture}
  \draw[-] (4, 0) node[below] {$x_t$} -- (4, 1);
  \draw[-] (6, 0) node[below] {$y_{t-1}$} -- (6, 1);
  \draw[-] (4, 1) -- (6, 1);
  \draw[-] (5, 1) -- (5, 1.5);
  \draw (4.5, 1.5) rectangle (5.5, 2.5);
  \draw (4.5, 1.5) .. controls (5, 1.5) and (5, 2.5) .. (5.5, 2.5);
  \draw (4.5, 2) node[left] {Nonlinear Transformation};
  \draw (5, 3) node[left] {$c_t$};
  \draw[-] (5, 2.5) -- (5, 4);

  \draw[-] (5, 4) node[circle,fill,inner sep=1pt,label=left:Input Gate]{} node[above right] {$i_t$} -- (6, 4);
  \draw (6, 3.5) rectangle (7, 4.5);
  \draw (6, 3.5) .. controls (6.5, 3.5) and (6.5, 4.5) .. (7, 4.5);
  \draw[-] (7, 4) -- (8, 4);
  \draw[-] (8, 5) -- (8, 3);
  \draw[-] (8, 3) -- (9, 3) node[right] {$x_t$};
  \draw[-] (8, 5) -- (9, 5) node[right] {$y_{t-1}$};

  \draw[-] (5, 4) -- (5, 6);
  \draw (4.5, 6) rectangle (5.5, 7);
  \draw[-] (4.5, 6) -- (5.5, 7);
  \draw (4.5, 6.5) node[above left] {Internal State};
  \draw (4.5, 6.5) node[below left] {$S_t = i_t \odot c_t + f_t \odot S_{t-1}$};
  \draw [-] (5, 7) .. controls (5, 7.5) and (6, 7.5) .. (6, 7);
  \draw [-] (6, 7) -- (6, 6);
  \draw [<-] (5, 6) .. controls (5, 5.5) and (6, 5.5) .. (6, 6);

  \draw[-] (6, 6.5) node[circle,fill,inner sep=1pt,label=below right:Forget Gate](Forget Gate){} node[above right] {$f_t$} -- (8.5, 6.5);
  \draw (8.5, 6) rectangle (9.5, 7);
  \draw (8.5, 6) .. controls (9, 6) and (9, 7) .. (9.5, 7);
  \draw[-] (9.5, 6.5) -- (10, 6.5);
  \draw[-] (10, 5.5) -- (10, 7.5);
  \draw[-] (10, 5.5) -- (10.5, 5.5) node[right] {$x_t$};
  \draw[-] (10, 7.5) -- (10.5, 7.5) node[right] {$y_{t-1}$};

  \draw (5, 7) -- (5, 8);
  \draw (4.5, 8) rectangle (5.5, 9);
  \draw (4.5, 8.5) node[left] {Nonlinear Transformation};
  \draw (4.5, 8) .. controls (5, 8) and (5, 9) .. (5.5, 9);
  \draw[->] (5, 9) -- (5, 11);

  \draw[-] (5, 10) node[circle,fill,inner sep=1pt,label=left:Output Gate](Output Gate){} node[above right] {$o_t$} -- (6, 10);
  \draw (6, 9.5) rectangle (7, 10.5);
  \draw (6, 9.5) .. controls (6.5, 9.5) and (6.5, 10.5) .. (7, 10.5);
  \draw[-] (7, 10) -- (8, 10);
  \draw[-] (8, 9) -- (8, 11);
  \draw[-] (8, 9) -- (9, 9) node[right] {$x_t$};
  \draw[-] (8, 11) -- (9, 11) node[right] {$y_{t-1}$};

\end{tikzpicture}
\caption{Diagram of the structure of a single LSTM cell}\label{LSTM_fig1}
\end{figure}
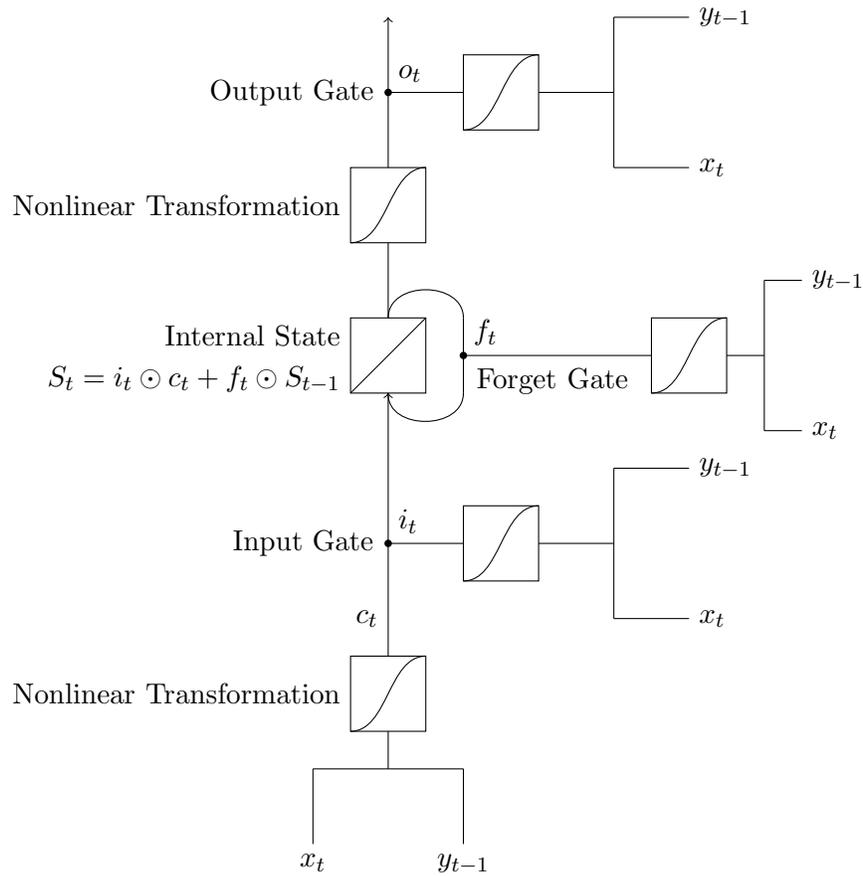

A layer of LSTM cells takes a sequence of vectors as input $x = (x_1, x_2, x_3,\ldots,x_T)$ and outputs a sequence of vectors $y = (y_1, y_2, y_3,\ldots,y_T)$. The output vectors are calculated by iterating through the following equations from $t = 1$ to $T$:

\[c_t = g(W_{cx} x_t + W_{cy} y_{t-1} + b_c)\]
\[i_t = \sigma(W_{ix} x_t + W_{iy} y_{t-1} + b_i)\]
\[f_t = \sigma(W_{fx} x_t + W_{fy} y_{t-1} + b_f)\]
\[o_t = \sigma(W_{ox} x_t + W_{oy} y_{t-1} + b_o)\]
\[S_t = i_t \odot c_t+ f_t \odot S_{t-1}\]
\[y_t = o_t \odot \phi(S_t)\]
\\
$\sigma(x)$ is defined as a hard sigmoid function which can output 0 and 1. This means that the gates can fully close or open.

\[ \sigma(x) =
  \begin{cases} 
    0 & x \leq -2.5 \\
    0.2x + 0.5 & -2.5 \leq x \leq 2.5 \\
    1 & 2.5 \geq x 
 \end{cases}
\]
\\
And

\[ \phi(x) = g(x) = tanh(x) \]

\section{Network Topology}
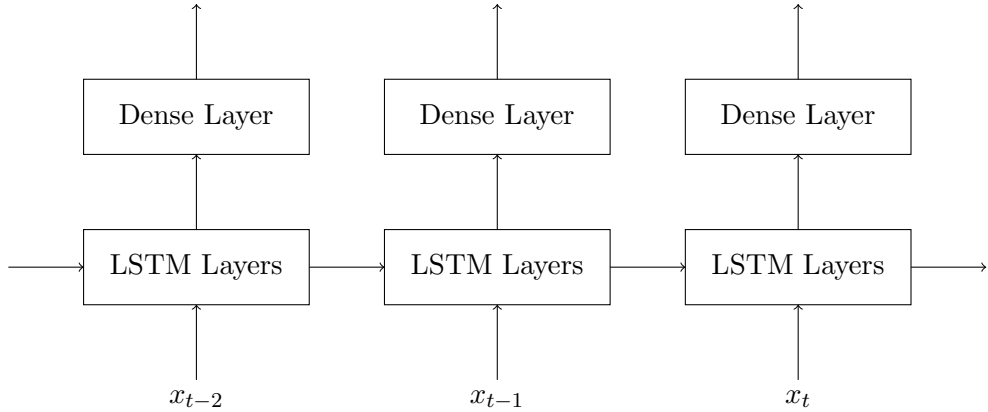
\begin{figure}[H]
\centering
\begin{tikzpicture}
  \draw[->] (2.5, 0) node[below] {$x_{t-2}$} -- (2.5, 1);
  \draw[->] (6.5, 0) node[below] {$x_{t-1}$} -- (6.5, 1);
  \draw[->] (10.5, 0) node[below] {$x_t$} -- (10.5, 1);

  \draw (1, 1) rectangle (4, 2);
  \draw (2.5, 1.5) node {LSTM Layers};
  \draw (5, 1) rectangle (8, 2);
  \draw (6.5, 1.5) node {LSTM Layers};
  \draw (9, 1) rectangle (12, 2);
  \draw (10.5, 1.5) node {LSTM Layers};

  \draw[->] (0, 1.5) -- (1, 1.5);
  \draw[->] (4, 1.5) -- (5, 1.5);
  \draw[->] (8, 1.5) -- (9, 1.5);
  \draw[->] (12, 1.5) -- (13, 1.5);

  \draw[->] (2.5, 2) -- (2.5, 3);
  \draw[->] (6.5, 2) -- (6.5, 3);
  \draw[->] (10.5, 2) -- (10.5, 3);

  \draw (1, 3) rectangle (4, 4);
  \draw (2.5, 3.5) node {Dense Layer};
  \draw (5, 3) rectangle (8, 4);
  \draw (6.5, 3.5) node {Dense Layer};
  \draw (9, 3) rectangle (12, 4);
  \draw (10.5, 3.5) node {Dense Layer};

  \draw[->] (2.5, 4) -- (2.5, 5);
  \draw[->] (6.5, 4) -- (6.5, 5);
  \draw[->] (10.5, 4) -- (10.5, 5);
\end{tikzpicture}
\caption{Network architecture}\label{Arch_fig2}
\end{figure}

LSTM layers are used to create a set of learned recurrent feature generators. A single dense layer is stacked on top to accumulate the outputs of the LSTM layers into the predictions of the network.

\section{Weight Initialisations}

The feed forward weights are initialised by sampling from the uniform distribution \cite{FFDiff}.

\[W \sim U \left[ -\frac{\sqrt{6}}{\sqrt{n_{in} + n_{out}}}, \frac{\sqrt{6}}{\sqrt{n_{in} + n_{out}}} \right] \]
\\

The recurrent weights are intialised by performing singular value decomposition on a 0 mean and unit variance Guassian distribution.

\[W = U S V^T\]
\\
This gives 2 random orthonormal matrices $U$ and $V$ and

\[W := U\]
\\
This results in recurrent weight matrices with unit maximal eigenvalue. As a result, the internal state does not explode as multiplying by the weight matrices repeatedly does not increase the Euclidean norm of the internal state vector.

Finally, the forget bias units are initialised to 1 and the other biases to 0.

\section{Feature Selection and Targets}

Google's daily stock data are used from January 1st 2005 to December 31st 2014 to create the training set and the stock data from January 1st 2015 to December 31st 2015 to create the test set. The data is obtained through the Yahoo finance API.

The data used for network inputs include the open, high, low, close and volume, and the network targets are open, high, low, close of the next day. The data and targets are both normalised through converting them into returns via percentage change.

\[ \hat{x_t} = \frac{x_t}{x_{t-1}} - 1 \]

This scales the values to be much smaller as stocks typically move little data to day, and also bounds the magnitude of the data. This makes it easier for the network to learn.

\section{Training Methodology}

We improve the training process by adding in a pretrainign step as follows. First, the full sequence of 2005 to 2014 is reduced to a list of sequences of length 2 via sliding windows. This presents a much easier problem for the LSTM to solve, thus the LSTM is trained on it for 1 epoch. The length is then doubled to 4 and sliding windows are applied to reduce the full sequence to a list of sequences of length 4. Training is applied in the same analogous way. This is repeated up to and including sequences of length 256. For sequences of length 256, training is done for 100 epochs with the same batchsize to complete the training process.

All training are done using the ADAM optimiser with default parameters and learning rate. The batchsize was maintained at 20 sequences per batch\cite{ADAM}.

\section{Experiments}

\begin {table}[H]
\centering
  \begin{tabular}{cc|c|c|c|c|}
	\cline{3-6}
    & & \multicolumn{4}{c|}{Hidden Layer Size} \\ \cline{3-6}
    & & 50 & 100 & 250 & 500 \\ \cline{1-6}

    \multicolumn{1}{ |c  }{\multirow{3}{*}{Hidden Layers} } &
      \multicolumn{1}{ |c| }{1} & 0.0154 & 0.0236 & 0.0139 & 0.0135 \\ \cline{2-6}
      \multicolumn{1}{ |c  }{} &
      \multicolumn{1}{ |c| }{2} & 0.0152 & 0.0166 & 0.0141 & 0.0152 \\ \cline{2-6}
      \multicolumn{1}{ |c  }{} &
      \multicolumn{1}{ |c| }{3} & 0.0141 & 0.0134 & 0.0105 & 0.0130 \\ \cline{2-6}
      \cline{1-6}
  \end{tabular}

\caption{Returns RMSE Of Specified Networks}
\end{table}

\section{Discussion}

As shown in Table 1, the returns data is generally resistant to overfitting. The network generally performs better when made deeper and wider with a few exceptions even when the number of parameters in the network are far greater than then number of unique training examples.

The results are also compared to the RMSE of a simple algorithm which predicts no change from day to day. The result of this algorithm is 0.0265 RMSE. Therefore the LSTM network is learning an effective pattern for prediction

\section*{Acknowledgements}

I would like to thank Alfie Howard for providing me with the necessary code to preprocess data. I would also like to thank Fran\c{c}ois Chollet for creating the Keras framework which was used to create all the necessary code for this paper.

\end{document}